\pdfoutput=1

\documentclass[11pt]{article}
\usepackage{amsmath}
\usepackage{multirow}
\usepackage{threeparttable}

\usepackage[preprint]{acl}

\usepackage{times}
\usepackage{latexsym}

\usepackage[T1]{fontenc}

\usepackage[utf8]{inputenc}

\usepackage{microtype}

\usepackage{inconsolata}

\usepackage{graphicx}

\title{Word2winners at SemEval-2025 Task 7: Multilingual and Crosslingual Fact-Checked Claim Retrieval}

\author{
 \textbf{Amirmohammad Azadi}\thanks{Equal contribution}, 
 \textbf{Sina Zamani}\footnotemark[1],
 \textbf{Mohammadmostafa Rostamkhani},
 \textbf{Sauleh Eetemadi}\\
 Iran University of Science and Technology \\
 \texttt{\{am\_azadi, sina\_zamani, mo\_rostamkhani97\}@comp.iust.ac.ir, sauleh@iust.ac.ir}
}

\begin{document}
\maketitle
\begin{abstract}
This paper describes our system for SemEval 2025 Task 7: Previously Fact-Checked Claim Retrieval. The task requires retrieving relevant fact-checks for a given input claim from the extensive, multilingual MultiClaim dataset, which comprises social media posts and fact-checks in several languages. To address this challenge, we first evaluated zero-shot performance using state-of-the-art English and multilingual retrieval models and then fine-tuned the most promising systems, leveraging machine translation to enhance crosslingual retrieval. Our best model achieved an accuracy of 85\% on crosslingual data and 92\% on monolingual data. \footnote{Our codes are available at \url{https://github.com/am-azadi/SemEval2025-Task7-Word2winners}}
\end{abstract}

\section{Introduction}

The spread of misinformation on social media poses a considerable challenge for fact-checkers, who must verify claims quickly and accurately, often across different languages. In our participation in SemEval 2025 Task 7 \citep{semeval2025task7}, we develop systems that retrieve the most relevant fact-checked claims for a given social media post, irrespective of linguistic differences.

The task is divided into two subtasks: 

\begin{itemize} 
    \item \textbf{Monolingual Retrieval:} In this subtask, both the social media posts and the fact-checks are in the same language. This setting allows us to focus on language-specific nuances and idiomatic expressions, thereby assessing the system's ability to match claims within a single language. 
    
    \item \textbf{Crosslingual Retrieval:} In this subtask, the social media post and the corresponding fact-checks are in different languages. This scenario reflects real-world situations where misinformation crosses language boundaries, necessitating the alignment of semantic representations across languages. 
\end{itemize}

The task uses a dataset derived from the original MultiClaim dataset \citep{Pikuliak_2023}, which includes over 150,000 fact-checks in 8 languages (English, Spanish, German, French, Arabic, Portuguese, Thai, and Malay) and social media posts in 14 languages.

Our approach begins with a preprocessing stage in which raw data is cleaned and summarized to reduce noise and standardize the content. Next, we employ transformer-based models—including LaBSE \citep{feng-etal-2022-language}, GTR-T5-Base \citep{ni2021largedualencodersgeneralizable}, and mE5 \citep{wang2024multilingual}—within a zero-shot evaluation framework to assess their ability to retrieve semantically similar claims across languages. Following the initial evaluation, the most effective models are fine-tuned using the task dataset to improve both precision and recall. An ensemble strategy based on majority voting is then applied to combine the outputs of these models, thereby mitigating the impact of individual model biases. This systematic approach offers a robust methodology for retrieving fact-checked claims in a multilingual context and addresses the challenges associated with misinformation detection.

Our experiments revealed that our ensemble approach consistently enhanced retrieval performance as measured by success-at-10 (S@10), showing an improvement of approximately 3–5 percentage points over the best single-model baseline. At the same time, a closer examination of the results indicates that the system tends to struggle with informal language, ambiguous expressions, and idiomatic usage, particularly in low-resource languages. These observations suggest that further refinement in preprocessing and model adaptation may help address the inherent variability and noise in social media content.

\section{Related Work}

Previously Fact-Checked Claim Retrieval (PFCR) is the task of retrieving the most relevant fact-checked claims from a dataset regarding a given social media post. Each claim in the dataset has been reviewed by experts and labeled as either misinformation or not. By ranking and retrieving the most relevant claims, we can infer a label for the post based on its similarity to existing claims.

For each social media post, the system is expected to retrieve 10 most related claims. The evaluation metric is Success@K which is defined as the proportion of cases where a relevant item appears within the top K results returned by a system. A model is considered successful if it includes the correct label within its top 10 ranked responses.

The benchmark includes early contributions from \citealp{kazemi-etal-2021-claim}, which supported a limited number of languages and primarily focused on specific topics, such as COVID-19. Their dataset originated from WhatsApp, containing 650 post-claim pairs. Despite its limitations, it established a growing research area, inspiring later datasets to expand and refine its scope.

For our study we used MultiClaim, the most comprehensive dataset in the benchmark. MultiClaim is multilingual, covering 39 languages and a wide range of topics while incorporating diverse cultural and social perspectives. It is collected from Facebook, Twitter, Instagram, and Google Fact Check Tools, ensuring broad coverage and diversity. With over 31,000 post-claim pairs, MultiClaim serves as an excellent resource for training models and enhancing fact-checking systems.

\begin{figure*}[t]
  \centering
  \includegraphics[width=0.80\linewidth]{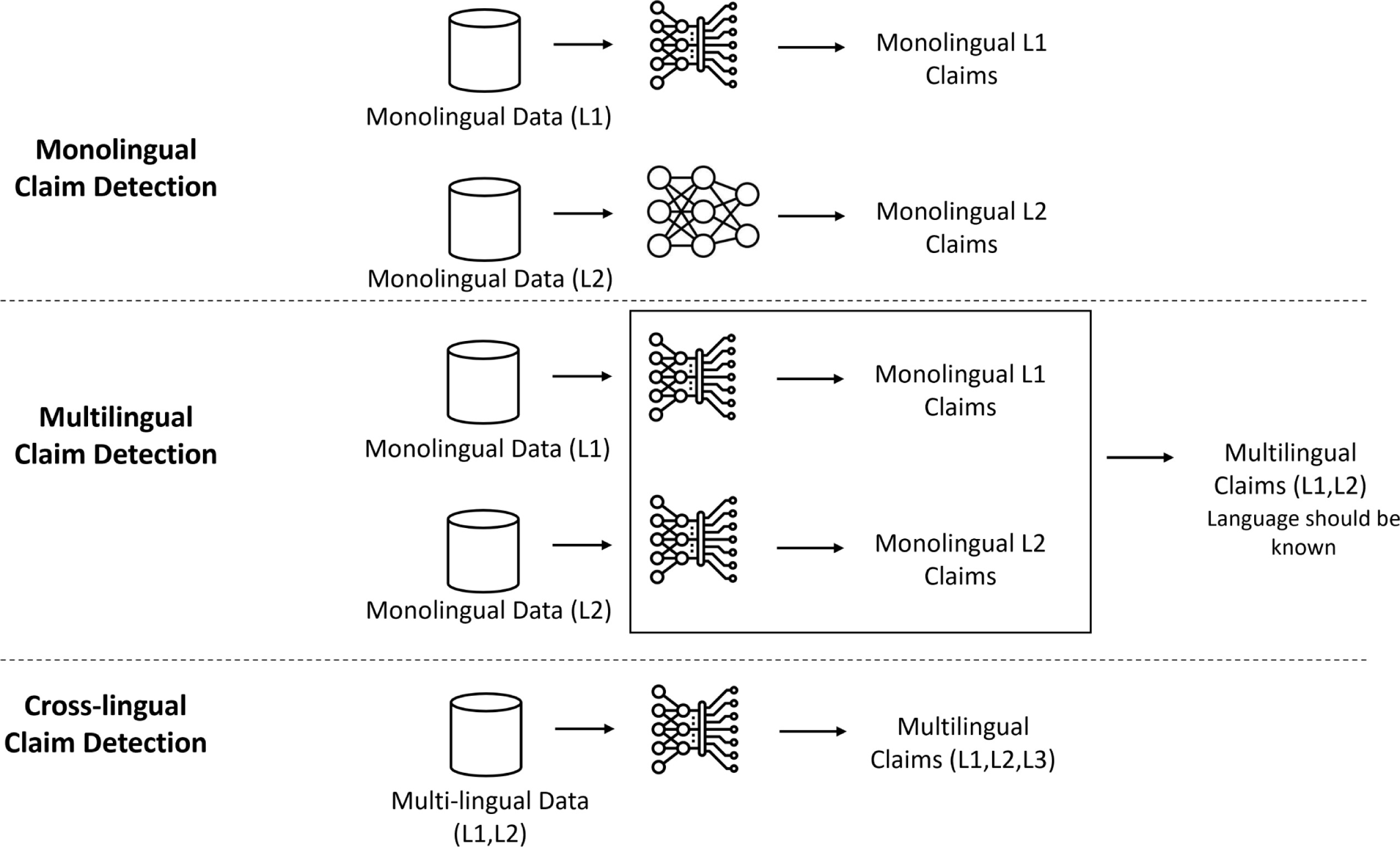}
  \caption{Types of information retrieval including monolingual, multilingual, and crosslingual retrieval illustrated by \citealp{PANCHENDRARAJAN2024100066}}
  \label{fig:claim_detection_types}
\end{figure*}

The SemEval task consists of two tracks: monolingual and crosslingual. In the monolingual setting, both the post and the claim are in the same language. In contrast, the crosslingual setting involves a post and a claim in different languages, introducing additional challenges for the system. The differences between these tracks and how their respective models operate are illustrated in Figure~\ref{fig:claim_detection_types}.

\section{System Overview}

\subsection{Preprocessing}
In the preprocessing stage, we first cleaned the raw social media data by removing irrelevant elements such as hashtags, emojis, and URLs, which often introduce noise and disrupt semantic analysis. Subsequently, for each post, we concatenated the OCR outputs with the original textual content to create a comprehensive representation. This combined content was then used in both the original language and its English translation, ensuring that subsequent multilingual retrieval tasks could effectively leverage the full scope of available information.

\subsection{Summarization}
To address the length and noisy nature of social media posts, we applied a summarization step to refine the content, remove irrelevant information, and improve its structure. To generate an effective summarization prompt, we used ReConcile Round Table \citep{chen2024reconcileroundtableconferenceimproves} involving three large language models—GPT-4o \citep{OpenAI_GPT4o_2024}, Claude 3.5 Sonnet \citep{Anthropic_Claude3.5_Sonnet_2024}, and LLaMA-3.3-70B-Instruct \citep{MetaAI_Llama3.3_70B_Instruct_2024}. Over three rounds, these models proposed different prompts, and a voting mechanism was used to select the most effective one. The final selected prompt was then provided to summarization models to generate concise and structured summaries, preserving key information while reducing noise.

\subsection{Zero-shot experiments}
To identify the most effective models for retrieving fact-checked claims, we conducted zero-shot evaluations using several state-of-the-art English and multilingual models. These models were selected based on their performance on established benchmarks such as MTEB \citep{muennighoff-etal-2023-mteb} and MMTEB \citep{enevoldsen2025mmtebmassivemultilingualtext}. By testing them on the training data, we aimed to assess their ability to capture semantic similarity between social media posts and fact-checks without task-specific fine-tuning. The results of these experiments provided insight into which models were best suited for multilingual claim retrieval and informed our selection of models for further fine-tuning.

\subsection{Fine-tuning}

Following the zero-shot experiments, we fine-tuned the most effective models using the training data. To construct the training inputs, each model was given a social media post paired with its corresponding fact-checked claim as a positive sample. The model was then trained using the Multiple Negatives Ranking Loss (MNRL), which optimizes the similarity between positive pairs while pushing negative samples further apart. This approach improves the model’s ability to distinguish relevant fact-checks from unrelated claims. The loss function is defined as follows:

\begin{equation}
\mathcal{L} = -\frac{1}{N} \sum_{i=1}^{N} \log \frac{e^{\text{sim}(q, p_i)}}{\sum_{j=1}^{N} e^{\text{sim}(q, n_j)}}
\end{equation}

where \( q \) represents the query (post), \( p_i \) is the positive sample, \( n_j \) are the negative samples, and sim denotes a similarity function such as cosine similarity.

\subsection{Training Configuration}

The following table outlines the hyperparameters used during the fine-tuning process:

\begin{table}[h]
    \centering
    \begin{tabular}{lc}
        \hline
        \textbf{Parameter} & \textbf{Value} \\
        \hline
        Batch Size & 2 - 16 \\  
        Learning Rate & 1e5 - 3e5 \\  
        Epochs & 1 - 3 \\  
        Warmup Steps & 100 \\  
        Hardware & (1 - 2) * T4 GPU \\
        Loss Function & MultipleNegativesRankingLoss \\  
        \hline
    \end{tabular}
    \caption{Training configuration for fine-tuning}
    \label{tab:training_config}
\end{table}

This fine-tuning process allowed the models to better capture the semantic relationships between social media posts and fact-checked claims, improving retrieval accuracy.

\subsection{Mojority Voting}

To leverage the strengths of the best-performing models, we applied a majority voting strategy to determine the most relevant fact-checks for each post. For every retrieved fact-check, we assigned a score based on two factors: the confidence of the model in selecting that fact-check and the model’s accuracy in the corresponding language. The final ranking was determined by summing the scores across all models, and the top 10 fact-checks with the highest scores were selected as the final output for each post. This approach aimed to balance the strengths of different models and improve retrieval robustness across languages.

\section{Results}

In the summarization phase, we employed several LLMs to reduce text length and eliminate noise. The models used included mT5-multilingual-XLSum \citep{hasan-etal-2021-xl}, BART-large-CNN \citep{DBLP:journals/corr/abs-1910-13461}, Falcon3-7B-Instruct \citep{Falcon3}, Qwen2.5-1.5B-Instruct, and Qwen2.5-7B-Instruct \citep{qwen2.5}, with the latter demonstrating the best performance. We used the last two models to summarize the input text, and the results are presented in Table~\ref{tab:summarization}. However, summarization significantly reduced model accuracy. This decline is due to the nature of the fact-checks dataset, which contains many similar instances without direct post-claim mappings in the pairs dataset. As a result, summarization increases the similarity between claims, making it harder for the model to generate distinct embeddings, thereby affecting similarity rankings.

\begin{table*}[]
\caption{The Impact of Summarization on the Zero-Shot Performance (S@10) of  Models}
\label{tab:summarization}
\centering
\resizebox{\textwidth}{!}{
\begin{tabular}{c|c|cccccccccc|}
\cline{2-12}
 & \textbf{Summarizer} & \textbf{fra} & \textbf{spa} & \textbf{eng} & \textbf{por} & \textbf{tha} & \textbf{deu} & \textbf{msa} & \textbf{ara} & \textbf{Crosslingual} & \textbf{Average} \\ \hline
\multicolumn{1}{|c|}{\multirow{3}{*}{SONAR}} & Raw & \textbf{0.65} & \textbf{0.60} & \textbf{0.62} & \textbf{0.63} & \textbf{0.40} & \textbf{0.45} & \textbf{0.61} & \textbf{0.59} & \textbf{0.62} & \textbf{0.57} \\
\multicolumn{1}{|c|}{} & Qwen2.5-1.5B-Instruct & 0.45 & 0.42 & 0.47 & 0.41 & 0.26 & 0.28 & 0.28 & 0.37 & 0.36 & 0.37 \\
\multicolumn{1}{|c|}{} & Qwen2.5-7B-Instruct & 0.62 & 0.59 & 0.59 & 0.57 & \textbf{0.40} & 0.37 & 0.50 & 0.54 & 0.48 & 0.53 \\ \hline
\multicolumn{1}{|c|}{\multirow{2}{*}{UAE-Large-V1}} & Raw & \textbf{0.85} & \textbf{0.81} & \textbf{0.75} & \textbf{0.82} & \textbf{0.93} & \textbf{0.66} & \textbf{0.85} & \textbf{0.79} & \textbf{0.62} & \textbf{0.81} \\
\multicolumn{1}{|c|}{} & Qwen2.5-1.5B-Instruct & 0.73 & 0.65 & 0.61 & 0.71 & 0.83 & 0.49 & 0.74 & 0.69 & 0.50 & 0.65 \\
\multicolumn{1}{|c|}{} & Qwen2.5-7B-Instruct & 0.77 & 0.79 & 0.65 & 0.74 & 0.89 & 0.58 & 0.80 & 0.73 & 0.55 & 0.74 \\ \hline
\end{tabular}
}
\end{table*}

A wide range of bi-encoder models can be used for information retrieval, each designed to extract embeddings from the input text. By computing the cosine similarity between embedding pairs, we can determine how similar they are. To identify the most effective models, we conducted zero-shot experiments with several candidates, testing multilingual models on the original dataset as well as its English translation. Interestingly, using the English translation typically lowers monolingual accuracy. This is because multilingual models are designed to generalize across multiple languages rather than being optimized for English specifically. However, translating the input text improves crosslingual accuracy. In crosslingual scenarios, where the post and claim are in different languages, multilingual models struggle to generate similar embeddings for semantically equivalent content across languages, leading to better differentiation after translation. We also tested English-specific models, which outperformed multilingual models. Since these models are trained exclusively on English data, they yield the best results when applied to the translated dataset. Overall, multilingual models offer better generalization across languages, while English models excel in single-language tasks. The results of our experiments are presented in Tables \ref{tab:zero-shot-multilingual} and \ref{tab:zero-shot-monolingual}.

\begin{table*}[]
\caption{Multilingual models' Zero-shot performance (S@10)}
\label{tab:zero-shot-multilingual}
\centering
\resizebox{\textwidth}{!}{
\begin{threeparttable}
\begin{tabular}{cc|cccccccccc|}
\cline{3-12}
 & & \textbf{fra} & \textbf{spa} & \textbf{eng} & \textbf{por} & \textbf{tha} & \textbf{deu} & \textbf{msa} & \textbf{ara} & \textbf{Crosslingual} & \textbf{Average} \\ \hline
\multicolumn{1}{|c|}{\multirow{2}{*}{GTE-Multilingual-Base}} & Original & 0.82 & 0.87 & 0.77 & 0.79 & \textbf{0.96} & 0.75 & \textbf{0.89} & \textbf{0.83} & 0.63 & \textbf{0.84} \\
\multicolumn{1}{|c|}{} & English & 0.86 & 0.86 & 0.77 & 0.81 & 0.91 & \textbf{0.78} & 0.88 & 0.79 & 0.66 & 0.83 \\ \hline
\multicolumn{1}{|c|}{\multirow{2}{*}{Multilingual-E5-Large}} & Original & 0.73 & 0.84 & 0.61 & 0.77 & 0.91 & 0.64 & 0.77 & 0.75 & 0.50 & 0.75 \\
\multicolumn{1}{|c|}{} & English & 0.73 & 0.79 & 0.66 & 0.74 & 0.86 & 0.68 & 0.79 & 0.53 & 0.46 & 0.72 \\ \hline
\multicolumn{1}{|c|}{\multirow{2}{*}{Multilingual-E5-Large-Instruct}} & Original & \textbf{0.87} & \textbf{0.90} & 0.76 & \textbf{0.84} & \textbf{0.96} & 0.64 & 0.85 & 0.70 & 0.62 & 0.81 \\
\multicolumn{1}{|c|}{} & English & 0.84 & 0.88 & 0.76 & 0.80 & \textbf{0.96} & 0.62 & 0.80 & 0.74 & \textbf{0.68} & 0.80 \\ \hline
\multicolumn{1}{|c|}{\multirow{2}{*}{KaLM-Embedding-Multilingual-mini-v1}} & Original & \textbf{0.87} & 0.88 & \textbf{0.79} & 0.79 & 0.93 & 0.72 & 0.82 & 0.81 & 0.56 & 0.83 \\
\multicolumn{1}{|c|}{} & English & \textbf{0.87} & 0.89 & \textbf{0.79} & 0.80 & 0.93 & 0.67 & 0.88 & 0.77 & 0.66 & 0.83 \\ \hline
\multicolumn{1}{|c|}{\multirow{2}{*}{LaBSE}} & Original & 0.75 & 0.65 & 0.51 & 0.68 & 0.86 & 0.52 & 0.71 & 0.69 & 0.39 & 0.68 \\
\multicolumn{1}{|c|}{} & English & 0.70 & 0.61 & 0.51 & 0.61 & 0.88 & 0.45 & 0.60 & 0.73 & 0.38 & 0.64 \\ \hline
\multicolumn{1}{|c|}{\multirow{2}{*}{Paraphrase-Multilingual-MPNet-Base-v2}} & Original & 0.76 & 0.62 & 0.60 & 0.58 & 0.93 & 0.46 & 0.73 & 0.60 & 0.39 & 0.66 \\
\multicolumn{1}{|c|}{} & English & 0.80 & 0.71 & 0.61 & 0.72 & 0.90 & 0.54 & 0.86 & 0.76 & 0.50 & 0.74 \\ \hline
\multicolumn{1}{|c|}{\multirow{1}{*}{SONAR}} & Original & 0.65 & 0.60 & 0.62 & 0.63 & 0.40 & 0.45 & 0.61 & 0.59 & 0.62 & 0.57 \\ \hline
\multicolumn{1}{|c|}{\multirow{1}{*}{BGE-M3}} & Original & 0.84 & 0.85 & 0.70 & 0.81 & 0.93 & 0.66 & 0.88 & 0.74 & 0.55 & 0.80 \\ \hline
\multicolumn{1}{|c|}{\multirow{1}{*}{XLM-R-100langs-BERT-Base-nli-stsb-mean-tokens}} & Original & 0.57 & 0.45 & 0.40 & 0.45 & 0.67 & 0.31 & 0.46 & 0.45 & 0.40 & 0.47 \\ \hline
\multicolumn{1}{|c|}{\multirow{1}{*}{GTR-T5-Base}} & Original & 0.76 & 0.66 & 0.70 & 0.67 & 0.18 & 0.57 & 0.52 & 0.08 & 0.33 & 0.52 \\ \hline
\multicolumn{1}{|c|}{\multirow{1}{*}{Sentence-T5-Base}} & Original & 0.40 & 0.44 & 0.52 & 0.48 & 0.16 & 0.29 & 0.32 & 0.08 & 0.21 & 0.34 \\ \hline
\end{tabular}
\end{threeparttable}
}
\end{table*}

\begin{table*}[]
    \caption{English models' Zero-shot performance (S@10)}
    \label{tab:zero-shot-monolingual}
    \centering
    \resizebox{\textwidth}{!}{
    \begin{threeparttable}
            \begin{tabular}{cc|cccccccccc|}
                \cline{3-12}
                 & & \textbf{fra} & \textbf{spa} & \textbf{eng} & \textbf{por} & \textbf{tha} & \textbf{deu} & \textbf{msa} & \textbf{ara} & \textbf{Crosslingual} & \textbf{Average} \\ \hline
                \multicolumn{1}{|c|}{\multirow{1}{*}{All-MPNet-Base-v2}} & English & 0.82 & 0.80 & 0.69 & 0.74 & 0.88 & 0.63 & 0.83 & 0.81 & 0.56 & 0.78 \\ \hline
                \multicolumn{1}{|c|}{\multirow{1}{*}{All-MiniLM-L6-v2}} & English & 0.83 & 0.80 & 0.68 & 0.74 & 0.90 & 0.64 & 0.83 & 0.79 & 0.55 & 0.78 \\ \hline
                \multicolumn{1}{|c|}{\multirow{1}{*}{Facebook-Contriever}} & English & 0.85 & 0.78 & 0.68 & 0.71 & 0.93 & 0.70 & 0.79 & 0.79 & 0.55 & 0.78 \\ \hline
                \multicolumn{1}{|c|}{\multirow{1}{*}{GTR-T5-Large}} & English & 0.83 & 0.83 & 0.73 & 0.80 & 0.88 & 0.69 & 0.79 & 0.79 & 0.60 & 0.80 \\ \hline
                \multicolumn{1}{|c|}{\multirow{1}{*}{Sentence-T5-Large}} & English & 0.64 & 0.61 & 0.53 & 0.58 & 0.88 & 0.30 & 0.74 & 0.62 & 0.39 & 0.62 \\ \hline
                \multicolumn{1}{|c|}{\multirow{1}{*}{MS Marco-BERT-Base-dot-v5}} & English & 0.85 & 0.83 & 0.72 & 0.81 & 0.95 & 0.65 & 0.86 & 0.79 & 0.60 & 0.81 \\ \hline
                \multicolumn{1}{|c|}{\multirow{1}{*}{UAE-Large-v1}} & English & 0.85 & 0.81 & 0.75 & 0.82 & 0.93 & 0.66 & 0.85 & 0.79 & 0.62 & 0.81 \\ \hline
                \multicolumn{1}{|c|}{\multirow{1}{*}{Bilingual-Embedding-Small}} & English & 0.88 & 0.87 & 0.78 & 0.81 & 0.96 & \textbf{0.74} & 0.89 & \textbf{0.83} & 0.65 & 0.85 \\ \hline
                \multicolumn{1}{|c|}{\multirow{1}{*}{Bilingual-Embedding-Large}} & English & \textbf{0.91} & \textbf{0.91} & \textbf{0.83} & \textbf{0.84} & \textbf{1.00} & \textbf{0.74} & \textbf{0.90} & 0.81 & \textbf{0.72} & \textbf{0.87} \\ \hline
                \multicolumn{1}{|c|}{\multirow{1}{*}{BGE-M3-custom-fr}} & English & 0.85 & 0.86 & 0.74 & 0.78 & 0.96 & 0.72 & 0.82 & 0.80 & 0.60 & 0.82 \\ \hline
            \end{tabular}
    \end{threeparttable}
    }
\end{table*}

\begin{figure}[]
  \centering
  \includegraphics[width=\columnwidth]{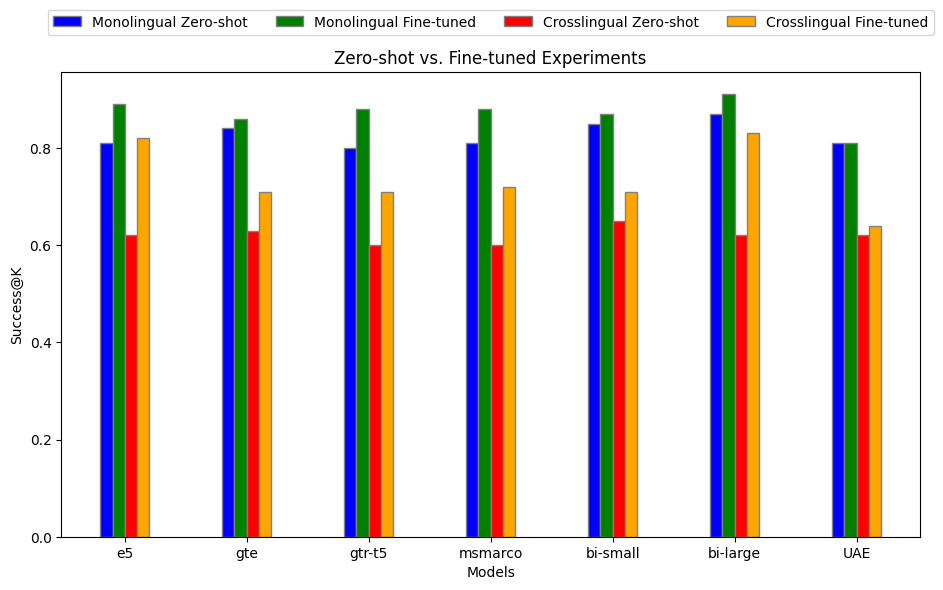}
  \caption{Zero-shot vs Fine-Tuning Performance (S@10)}
  \label{fig:zero-shot-vs-fine-tuning}
\end{figure}

After conducting multiple zero-shot experiments, we selected the best-performing models and fine-tuned them on the training dataset. The multilingual models were trained on the various languages present in the dataset, while the monolingual models were trained on its English translation. The results are shown in Tables \ref{tab:fine-tuned-multilingual} and \ref{tab:fine-tuned-monolingual}. Fine-tuning significantly improved model performance, highlighting its role in adapting to dataset-specific patterns. Figure \ref{fig:zero-shot-vs-fine-tuning} illustrates this effect in both monolingual and crosslingual settings, demonstrating substantial improvements across models—except for the UAE-Large-V1 model, which overfit rapidly and failed to generalize to test data.

\begin{figure}[]
  \centering
  \includegraphics[width=\columnwidth]{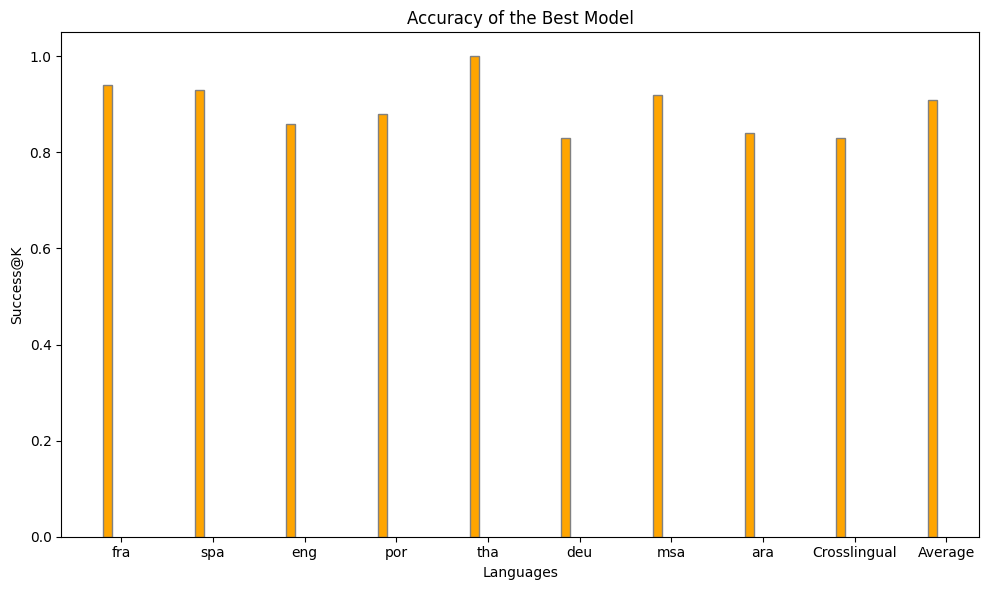}
  \caption{The best model's performance (S@10) on different languages}
  \label{fig:languages}
\end{figure}

Notably, fine-tuning benefited the crosslingual setting more than the monolingual one. This is because multilingual models, not being explicitly optimized for information retrieval tasks, rely on fine-tuning to better learn cross-language mappings. Figure \ref{fig:languages} further reveals accuracy discrepancies across languages. English, having the largest fact-checking portion in the dataset, presents greater difficulty in retrieving exact matches, leading to lower accuracy. In contrast, languages with smaller fact-checking portions, such as Thai, achieve higher accuracy due to the availability of more distinct examples. Additionally, some languages like Arabic, suffer from lower accuracy due to limited training data and reduced model familiarity.

\begin{table*}[]
\caption{Multilingual models' performance (S@10)}
\label{tab:fine-tuned-multilingual}
\centering
\resizebox{\textwidth}{!}{
\begin{threeparttable}
\begin{tabular}{cc|cccccccccc|}
\cline{3-12}
 & & \textbf{fra} & \textbf{spa} & \textbf{eng} & \textbf{por} & \textbf{tha} & \textbf{deu} & \textbf{msa} & \textbf{ara} & \textbf{Crosslingual} & \textbf{Average} \\ \hline
\multicolumn{1}{|c|}{\multirow{1}{*}{Baseline}} & Original & - & - & - & - & - & - & - & - & 0.22 & 0.70 \\ \hline
\multicolumn{1}{|c|}{\multirow{1}{*}{Multilingual-E5-Large-Instruct}} & Original & \textbf{0.93} & \textbf{0.92} & \textbf{0.82} & \textbf{0.89} & \textbf{0.98} & \textbf{0.83} & \textbf{0.90} & \textbf{0.85} & \textbf{0.82} & \textbf{0.89} \\ \hline
\multicolumn{1}{|c|}{\multirow{1}{*}{GTE-Multilingual-Base}} & Original & 0.85 & 0.90 & 0.80 & 0.82 & \textbf{0.98} & 0.81 & \textbf{0.90} & 0.83 & 0.71 & 0.86 \\ \hline
\end{tabular}
\end{threeparttable}
}
\end{table*}

\begin{table*}[]
\caption{English models' performance (S@10)}
\label{tab:fine-tuned-monolingual}
\centering
\resizebox{\textwidth}{!}{
\begin{threeparttable}
\begin{tabular}{cc|cccccccccc|}
\cline{3-12}
 & & \textbf{fra} & \textbf{spa} & \textbf{eng} & \textbf{por} & \textbf{tha} & \textbf{deu} & \textbf{msa} & \textbf{ara} & \textbf{Crosslingual} & \textbf{Average} \\ \hline
\multicolumn{1}{|c|}{\multirow{1}{*}{Baseline}} & English & - & - & - & - & - & - & - & - & 0.56 & 0.83 \\ \hline
\multicolumn{1}{|c|}{\multirow{1}{*}{GTR-T5-Large}} & English & 0.91 & 0.90 & 0.81 & 0.86 & \textbf{1.00} & 0.76 & 0.90 & \textbf{0.85} & 0.71 & 0.88 \\ \hline
\multicolumn{1}{|c|}{\multirow{1}{*}{MS Marco-BERT-Base-dot-v5}} & English & 0.89 & 0.90 & 0.80 & 0.86 & \textbf{1.00} & 0.81 & \textbf{0.92} & 0.81 & 0.72 & 0.88 \\ \hline
\multicolumn{1}{|c|}{\multirow{1}{*}{Bilingual-Embedding-Small}} & English & 0.89 & 0.89 & 0.79 & \textbf{0.88} & 0.98 & 0.78 & 0.91 & 0.80 & 0.71 & 0.87 \\ \hline
\multicolumn{1}{|c|}{\multirow{1}{*}{Bilingual-Embedding-Large}} & English & \textbf{0.94} & \textbf{0.93} & \textbf{0.86} & \textbf{0.88} & \textbf{1.00} & \textbf{0.83} & \textbf{0.92} & 0.84 & \textbf{0.83} & \textbf{0.91} \\ \hline
\multicolumn{1}{|c|}{\multirow{1}{*}{UAE-Large-v1}} & English & 0.86 & 0.86 & 0.73 & 0.82 & 0.96 & 0.67 & 0.85 & 0.75 & 0.64 & 0.81 \\ \hline
\end{tabular}
\end{threeparttable}
}
\end{table*}

While a single model may achieve the highest overall performance, it does not necessarily produce the best results across all languages and scenarios. To address this, a voting mechanism can be employed to leverage the strengths of multiple models, leading to more robust and accurate predictions. The results of this approach are presented in Table \ref{tab:MajorityVote}.

\begin{table*}[]
\caption{Majority voting performance (S@10) compared to the best model}
\label{tab:MajorityVote}
\centering
\resizebox{\textwidth}{!}{
\begin{tabular}{c|cccccccccc|}
\cline{2-11}
 & \textbf{fra} & \textbf{spa} & \textbf{eng}  & \textbf{por} & \textbf{tha} & \textbf{deu} & \textbf{msa} & \textbf{ara} & \textbf{Crosslingual} & \textbf{Overall} \\ \hline
\multicolumn{1}{|c|}{\multirow{1}{*}{Majority voting}} & \textbf{0.93} & \textbf{0.94} & \textbf{0.85} & \textbf{0.92} & \textbf{1.00} & \textbf{0.90} & \textbf{0.94} & \textbf{0.85} & \textbf{0.85} & \textbf{0.92} \\ \hline
\multicolumn{1}{|c|}{\multirow{1}{*}{Bilingual-Embedding-Large}} & 0.94 & 0.93 & 0.86 & 0.88 & \textbf{1.00} & 0.83 & 0.92 & 0.84 & 0.83 & 0.91 \\ \hline
\end{tabular}
}
\end{table*}

\section{Conclusion}
In this study, we developed a system for retrieving fact-checked claims from a multilingual dataset. The system uses several stages, including data preprocessing, summarization, zero-shot evaluation, fine-tuning, and an ensemble majority voting approach to match social media posts with relevant fact-checks. Our experiments show that fine-tuning improves performance, especially in crosslingual settings when combined with machine translation. The ensemble strategy also helps to overcome the limitations of individual models, leading to high accuracy in both crosslingual and monolingual tasks.

However, the results reveal some challenges, such as managing informal language and ambiguous expressions, particularly in low-resource languages. Future work should focus on enhancing preprocessing methods, exploring alternative summarization techniques, and incorporating more language resources to improve overall system robustness. Overall, this study provides a clear and effective framework for fact-checked claim retrieval, which is essential for addressing the spread of misinformation.

\bibliography{custom}

\end{document}